\documentclass[number]{autarticle}
\usepackage[utf8]{inputenc}
\usepackage{amsmath}
\usepackage{hyperref}
\usepackage{etoolbox}
\usepackage{mathtools, thmtools}
\usepackage{amsfonts}
\usepackage{amssymb}
\usepackage{graphicx}
\usepackage{caption , subcaption}
\usepackage{lipsum}
\usepackage{fancyheadings}

\hypersetup {hidelinks}
\DeclareMathOperator*{\argmin}{argmin}

\begin{document}
		\begin{frontmatter}
			
			\title{Lipschitzness Effect of a Loss Function on Generalization Performance of Deep Neural Networks Trained by Adam and AdamW Optimizers}
			
			
			\author[]{Mohammad Lashkari\footnotemark[1]{}}{*}
			\author[]{Amin Gheibi\footnotemark[1]{}}{}
			\ead{mohammad.l@aut.ac.ir, amin.gheibi@aut.ac.ir}
			
			\address[add1]{Mathematics and Computer Science Department, Amirkabir University of Technology, Tehran, Iran}
			
			\begin{abstract}
				The generalization performance of deep neural networks with regard to the optimization algorithm is one of the major concerns in machine learning. This performance can be affected by various factors. In this paper, we theoretically prove that the Lipschitz constant of a loss function is an important factor to diminish the generalization error of the output model obtained by Adam or AdamW. The results can be used as a guideline for choosing the loss function when the optimization algorithm is Adam or AdamW.
				
				In addition, to evaluate the theoretical bound in a practical setting, we choose the human age estimation problem in computer vision. For assessing the generalization better, the training and test datasets are drawn from different distributions. Our experimental evaluation shows that the loss function with a lower Lipschitz constant and maximum value improves the generalization of the model trained by Adam or AdamW.
			\end{abstract}

		
			
		\end{frontmatter}

	\section{Introduction}
	The adaptive moment estimation (Adam) algorithm is one of the most widely used optimizers for training deep learning models. Adam is an efficient algorithm for stochastic optimization, based on adaptive estimates of first-order and second-order moments of gradient \cite{kingma2014adam}. The method is computationally efficient and has little memory usage. Adam is much more stable than stochastic gradient descent (SGD) and the experiments of work \cite{kingma2014adam} show that it is faster than previous stabilized versions of SGD, such as SGDNesterov \cite{hazan2015beyond}, RMSProp \cite{tieleman2012lecture} and AdaGrad \cite{duchi2011adaptive} to minimize the loss function in the training phase. It is recently used in several machine learning problems and performs well. Thus, any improvement in the generalization performance of a model trained by Adam is essential.
	
	One of the main concerns in machine learning is the generalization performance of deep neural networks (DNNs). A generalization measurement criterion is the generalization error which is defined as the difference between the true risk and the empirical risk of the output model \cite{akbari2021does}. One established way to address the generalization error of machine learning models in order to derive an upper bound for it, is the notion of uniform stability \cite{akbari2021does,bousquet2002stability, hardt2016train}. Roughly speaking, the uniform stability measures the difference in the error of the output model caused by a slight change in the training set. The pioneering work of \cite{bousquet2002stability}, shows that if a deterministic learning algorithm is more stable, then the generalization error of the ultimate model achieves a tighter upper bound. In the following work of \cite{hardt2016train}, Hardt \textit{et al.} extend the notion of uniform stability to randomized learning algorithms to drive an upper bound for the expected generalization error of a DNN trained by SGD. They prove that SGD is more stable, provided that the number of iterations is sufficiently small.
	In the recent work of \cite{akbari2021does}, Ali Akbari \textit{et al.} derive a high probability generalization error bound instead of an expected generalization error bound. They demonstrate that if SGD is more uniformly stable, then the generalization error bound is tighter. They also proved the direct relationship between the uniform stability of SGD and loss function properties i.e. its Lipschitzness, resulting in the generalization error connection to the Lipschitz constant and the maximum value of a loss function.
	
	In this paper, we make the following contributions. First, we distinguish the relationship between the uniform stability of Adam and Lipschitzness of a loss function. Second, we derive an upper bound for the generalization error of a DNN trained by Adam which is directly related to the Lipschitz constant and the maximum value of a loss function. Third, we assess the uniform stability of AdamW optimizer which decouples weight decay from estimates of moments to make the regularization technique more effective \cite{loshchilov2017decoupled}. Fourth, we connect the generalization error of a DNN trained by AdamW to Lipschitzness and the maximum value of a loss function. In Experiments, we evaluate our theoretical results in the human age estimation problem.
	
	Human age estimation is one of the most significant topics in a wide variety of applications such as age-specific advertising, customer profiling, or recommending new things. However, we are facing many challenges to solve this problem. Face makeup, insufficient light, skin color, and unique features of each person are the factors that can affect the accuracy of the model. Based on these reasons, collecting more data cannot necessarily reduce the generalization error of the final models in this problem. Hence, it is a de facto example of label distribution learning where we can evaluate our theoretical bounds precisely as the authors of \cite{akbari2021does} did because it is a challenging problem, and if our theoretical analyzes are correct and advantageous, then we can enhance age estimation models. Furthermore, there are available age estimation datasets drawn from different distributions which gives us the ability to assess the generalization error of the models well. The practical results show that choosing a stable loss function based on our theoretical bounds, can improve the accuracy  and enhance the generalization performance of a model trained by Adam or AdamW.
	
	\section{Related Work}
	There is a variety of approaches to derive upper bounds for the generalization error including algorithmic stability \cite{akbari2021does, bousquet2002stability, hardt2016train, shalev2010learnability, banerjee22a, jakubovitz2018generalization},
	robustness \cite{ren2018learning, zahavy2016ensemble},
	PAC-Bayesian Theory \cite{neyshabur2018pac, guan22b}, and Vapnik-Chervonenkis
	(VC) dimension \cite{scarselli2018vapnik, basu2018deep, harvey2017nearly}. Each of these approaches theoretically analyzes some effective factors and gives the researchers some information which can enhance the generalization performance of deep learning models. 
	
	In the robustness theory, the generalization performance of models is measured according to the structure of models and how to use training data. In the work of \cite{ren2018learning}, a meta-learning algorithm was designed to learn what weights assign to training samples based on their gradient direction to improve the generalization ability. The algorithm can be applied to any type of neural network; it does not need hyper-parameter tuning and has good accuracy on unbalanced data. In the work of \cite{zahavy2016ensemble}, Zahavy \textit{et al.} defined a new measure called ensemble robustness, which measures the robustness of several hypotheses. Leveraging it, they demonstrated that a deep learning model can generalize well if its variance is controlled.
	
	The PAC-Bayesian theory is an approach to analyze Bayesian learning algorithms in which a prior distribution is considered on the hypothesis space and the output of the algorithm is also a distribution over this space. In the work of \cite{neyshabur2018pac}, an upper bound in terms of the product of the spectral norm of the layers and the Frobenius norm of the weights was derived for the generalization error of a neural network. Guan and Zhiwu in \cite{guan2022fast} generalized two generalization bounds i.e. kl-bound and Catoni-bound to meta-learning framework and proposed two classification algorithms with fast convergence rates by minimizing the aforementioned upper bounds.
	
	The VC-dimension is a measure of complexity, flexibility, and generalization of classification models created based on feed-forward neural networks and recurrent neural networks (RNNs). In the work of  \cite{scarselli2018vapnik}, this concept was extended to graph neural networks (GNNs) and recursive neural networks (RecNNs). The main finding of the work is that the upper bound for the VC-dimension of GNNs and RecNNs is comparable with the VC-dimension upper bound of RNNs. Also, the generalization ability of these models is directly related to the number of connected nodes \cite{scarselli2018vapnik}. The authors of \cite{basu2018deep}, derived an upper bound for the VC-dimension of convolutional neural networks, resulting in reducing the error of text classification models and increasing their generalization ability. In the work of \cite{harvey2017nearly}, an upper bound and a lower bound for the VC-dimension of neural networks with activation function $ ReLU $  are derived that both are polynomial in terms of the number of layers and the number of weights. All the results of the work can be generalized to any model with arbitrary piece-wise linear activation functions.
	
	We follow the notion of uniform stability in this paper. This notion was firstly introduced in \cite{bousquet2002stability} for deterministic algorithms. It was extended to randomized algorithms in the work of \cite{hardt2016train} to derive an expected upper bound for the generalization error which is directly related to the number of training epochs of SGD.
	
	Recently, based on the uniform stability definition of SGD, the generalization error of a DNN trained by it, with high probability, is upper-bounded by a vanishing function which is directly related to the Lipschitz constant and the maximum value of a loss function \cite{akbari2021does}. In our work, we analyze the uniform stability for Adam and AdamW and its relationship with the Lipschitz constant of a loss function. Then, we demonstrate how the characteristics of a loss function i.e. its Lipschitz constant and maximum value can be effective on the generalization performance of a DNN. We show that the loss function proposed in \cite{akbari2021does} to stabilize the training process when the optimizer is SGD, can also stabilize the training process and reduces the generalization error when the optimizer is Adam or AdamW. Other researchers can use our theoretical results to find new loss functions for training DNNs by Adam or AdamW.
	
	\section{Preliminaries}\label{prelim}
	Let $ X $ and $ Y \subseteq \mathbb{R}^\mathrm{M} $ be the input and output spaces of a problem respectively, and $ F $ be the set of all mappings from $ X $ to $ Y $. A learning problem is to find $ f^\theta: X \rightarrow Y$, parameterized by $ \theta \in H$ where $H \subset \mathbb{R}^K$ is a bounded set, containing all possible values for the neural network parameters \footnote{We know the number of iterations in the training phase is finite. Therefore the set of visible values for parameters in the training stage is finite. So we can assume the set of all possible values is an infinite bounded superset of visible values. We will the reason in Subsection \ref{adamw}}. Assume
	$\ell: Y \times Y \rightarrow \mathbb{R^+}$ denotes the loss function of the problem. The goal of a learning algorithm is to minimize the true risk $R_{true}(f^\theta) \coloneqq \mathbb{E}_{(\mathrm{x}, \mathrm{y})}\left[ \ell(f^\theta(\mathrm{x}) ,\mathrm{y}) \right]$ where $(\mathrm{x},\mathrm{y}) \in X \times Y$:
	\begin{equation}\label{learningprob}
	f^{\theta}_{true} = \argmin_{f^\theta \in F} R_{true}(f^\theta).
	\end{equation}
	Since the distribution of $X \times Y$ is unknown; $ f^\theta_{true} $ cannot be found in the equation \eqref{learningprob}. Hence, we have to estimate the true risk. Let $S \in (X \times Y)^N$ be the training set. The true risk is estimated by the empirical risk $ R_{emp}(f^\theta) \coloneqq \frac1N \sum_{i=1}^{N} \ell(f^\theta(\mathrm{x}_i), \mathrm{y}_i) $
	in which, $ N=|S| $ and $ (\mathrm{x}_i,\mathrm{y}_i) \in S $.
	In current deep learning algorithms, training the model means minimizing $ R_{emp}(f^\theta) $.
	In the rest of this paper, in the theorems and proofs, the loss function is denoted by
	$\ell(\hat{\mathrm{y}},\mathrm{y})$ where $ \hat{\mathrm{y}} $ is the predicted vector and $ \mathrm{y} $ is the target vector.
	\begin{Definition}[Partition] \label{partition}
		Suppose that $ S $ is a training set of size $ N $. Let $ 1 < k < N $ be a number that $ N $ is divisible by $ k $ (if it is not possible, we repeat a sample enough to make divisibility  possible). A partition of $ S $, which we denote by $ B_S = \lbrace B_1, B_2, \ldots, B_k \rbrace$, is a set of $ k $ subsets of $ S $ such that every sample is in exactly one of these subsets and the size of each subset is $\frac{N}{k}$.
	\end{Definition}
	We use Definition \ref{partition} to formalize the training process of deep learning models mathematically. Assume $S$ is the training set and $B_S = \lbrace B_1, B_2, \ldots, B_k \rbrace$ is a partition of it. Each element of $B_S$ represents a mini-batch of $S$. Without loss of generality we suppose that in each iteration of the optimization algorithm, a mini-batch $B_i \in B_S$ is randomly selected to the parameters be updated \footnote{The index of the mini-batch is randomly selected from $\lbrace1,2,\ldots,k\rbrace$.}. This is done by the algorithm using a random sequence $ R=\left(r_1,r_2,\ldots, r_T \right) $ of indices of elements in $ B_S $, where $ T $ is the number of iterations. We use $ f^\theta_{B_S,R} $ to denote the output model of the optimization algorithm,  applied to a partition $B_S$ and a random sequence $R$.
	
	\begin{Definition}[Generalization Error]
		Given a partition $B_S$ of a training set $S$ and a sequence $R$ of random indices of $B_S$ elements, the generalization error of $ f^\theta_{B_S,R} $ trained by an arbitrary optimization algorithm, is defined as $ E(f^\theta_{B_S,R}) = R_{true}(f^\theta_{B_S,R}) - R_{emp}(f^\theta_{B_S,R}). $
	\end{Definition}

	\begin{Definition}[Lipschitzness]
		Let $ Y \subseteq \mathbb{R}^\mathrm{M} $ be the output space of a problem. A loss function $\ell(\hat{\mathrm{y}},\mathrm{y})$ is $ \gamma $-Lipschitz with regard to its first argument, if $\, \forall \, \mathrm{y_1, y_2} \in Y$, we have:
		\[
		\mathrm{|\ell(y_1,y) - \ell(y_2,y)|} \leq \gamma \left\| \mathrm{y}_1-\mathrm{y}_2 \right\|,
		\]
		where $ \left\|.\right\| $ is the $L_2$ norm.
	\end{Definition}
	
	As mentioned before, uniform stability of the optimization algorithm is effective on the generalization performance of the ultimate model $ f^\theta_{B_S,R} $ \cite{akbari2021does}.
	We follow the uniform stability definition of work \cite{hardt2016train} to link Lipschitzness of the loss function to the generalization error of $ f^\theta_{B_S,R} $. For simplicity, moving forward, we denote $ f^\theta_{B_S,R} $ by $ f_{B_S,R} $  and $ E(f^\theta_{B_S,R}) $ by $ E(f_{B_S,R}) $.
	
	Along with the notion of uniform stability which we define in Section \ref{uniformsec}, another concept called bounded difference condition (BDC) affects the generalization error \cite{akbari2021does}:
	
	\begin{Definition}[BDC]\label{BDC}
		Consider two numbers $ k, T \in \mathbb{N}$. If $G: \{1,2,\ldots,k\}^T \rightarrow \mathbb{R^+}$, is a measurable function and for $R,R' \in Dom(G)$
		which are different only in two elements, constant $\rho$ exists such that \[
		\sup_{R,R'} |G(R')-G(R)| \leq \rho,
		\]
		then, $G(.)$ holds bounded difference condition (BDC) with the constant $ \rho $. We use the $\rho$-BDC expression to denote that a function holds this condition with the constant $\rho$.
	\end{Definition}

	In Definition \ref{BDC}, we assumed the slight change in the input to be the difference in two elements, which we will see its reason in the proof of the theorems. Intuitively, if a function satisfies the above condition, its value does not differ much due to a slight change in the input. Such functions are dense around their expectation with respect to the input random sequence $ R $ \cite{mcdiarmid1989method}.
	
	\section{Formulation of Age Estimation Problem} \label{problem}
	Our problem in the experimental part is human age estimation. Let $(\mathrm{x}, y)$ be a training sample where $ \mathrm{x} $ is the input image of a person's face and $y \in \mathbb{N} $ is the corresponding age label. Due to the correlation of the neighboring ages, classification methods based on single-label learning \cite{rothe2018deep} are not efficient because these methods ignore this correlation. Also, regression-based models are not stable to solve this problem \cite{akbari2021does}.
	
	According to the aforementioned reasons, another method based on label distribution learning (LDL) framework which was firstly introduced in the work of \cite{geng2016label}, is used for this problem \cite{akbari2021does}. In this method $ y $ is replaced by $\mathrm{y}=\left[ y_1, y_2,\ldots, y_\mathrm{M} \right] \in \mathbb{R}^\mathrm{M}$ where $ y_i $ is the probability of facial image $ \mathrm{x} $ belongs to class $ i $. As usual, $\mathrm{y}$ is assumed to be a normal distribution, centering at $y$ and standard deviation $\sigma$ which controls the spread of the distribution \cite{geng2016label}. Therefore, the output space, $Y$ is a subset of $\mathbb{R}^\mathrm{M}$ and our objective is to find $f^\theta$ which maps $\mathrm{x}$ to $\mathrm{y} \in Y$.
	\subsection{Loss Functions for Age Estimation Problem}
	Let $(\mathrm{x},\mathrm{y}) \in S$ be a training instance where $\mathrm{x}$ represents the facial image and $\mathrm{y} \in \mathbb{R}^\mathrm{M}$ is the corresponding label distribution. Consider $\hat{\mathrm{y}} = f^{\theta}(\mathrm{x})$, representing the estimated label distribution by $ f^{\theta} $. To obtain $ f^{\theta} $, a convex loss function named Kullback-Leibler (KL) divergence
	has been widely utilized. The KL loss function is defined as below:
	\[ \ell_{KL}(\hat{\mathrm{y}}, \mathrm{y}) = \sum_{m=1}^{\mathrm{M}} y_m \log(\frac{y_m}{ \hat{y}_m}). \]
	As an alternative to KL, another convex loss function called Generalized Jeffries-Matusita (GJM) distance has been proposed in \cite{akbari2021does} under the LDL framework, defined as
	
	\[ \ell_{GJM}(\hat{\mathrm{y}}, \mathrm{y}) = \sum_{m=1}^{\mathrm{M}} y_m \left| 1-\left(\frac{\hat{y}_m}{y_m}\right)^\alpha \right|^{\frac{1}{\alpha}}, \]
	
	where $\alpha \in (0,1]$. According to the experiments of \cite{akbari2021does}, the best value of $\alpha$ for good generalization is $0.5$. It has been proved that if $\alpha=0.5$, then the Lipschitz constant and the maximum value of GJM are less than the Lipschitz constant and the maximum value of KL respectively \footnote{It should be mentioned here that KL is not Lipschitz because when $x \rightarrow 0$ the derivative of $ \log(x) $ tends to infinity. So we have to bound its domain from left e.g. $(10^{-10}, +\infty)$ to make it Lipschitz. In contrast to KL, GJM does not have this issue.} \cite{akbari2021does}.
	
	\section{Uniform Stability and Generalization Error Analysis} \label{uniformsec}
	The notion of uniform stability was firstly introduced in \cite{bousquet2002stability} for deterministic learning algorithms. They demonstrate that smaller stability measure of the learning algorithm, the tighter generalization error is. However, their stability measure is limited to deterministic algorithms and is not appropriate for randomized learning algorithms such as Adam. Therefore, we follow \cite{akbari2021does, hardt2016train} to define the uniform stability measure for  randomized optimization algorithms generally:  
	\begin{Definition}[Uniform Stability] \label{uniformstabiliydef}
		Let $S$ and $S'$ denote two training sets drawn from a distribution $\mathbb{P}$. Suppose that $B_S$ and $B_{S'}$ of equal size k, are two partitions of $S$ and $S'$ respectively, which are different in only one element (mini-batch). Consider a random sequence $ R $ of  $\lbrace 1,2, \ldots, k \rbrace$ to select a mini-batch at each iteration of an optimization algorithm, $A_{opt}$. If $f_{B_S,R}$ and $f_{B_{S'},R}$ are output models obtained by $A_{opt}$ with the same initialization, then $A_{opt}$ is $\beta$-uniformly stable with regard to a loss function $ \ell $, if
		\[
		\forall S,S' \;\; \sup_{\mathrm{(x,y)}} \mathbb{E}_R \left[ |\ell(f_{B_{S'},R}(\mathrm{x}),\mathrm{y}) - \ell(f_{B_S,R}(\mathrm{x}),\mathrm{y})| \right] \leq \beta.
		\]
	\end{Definition}
	
	To evaluate the uniform stability of Adam and AdamW in order to prove its link to loss function properties, a lemma named \textbf{Growth recursion} which has been stated in \cite{hardt2016train} for SGD is central to our analysis. In the following, we state this lemma for an arbitrary iterative optimization algorithm, but before stating the lemma, we need some definitions. As we know, gradient-based optimization algorithms are iterative, and in each iteration, the network parameters are updated. Let $H$ be the set of all possible values for the neural network parameters. Let $A_{opt}$ be an arbitrary iterative optimization algorithm that runs $ T $ iterations. In the $ t $-th iteration, the update that is computed in the last command of the loop for the network parameters, is a function $A^t:H \rightarrow H$  mapping $\theta_{t-1}$ to $\theta_{t}$ for each $1 \leq t \leq T$. We call $A^t$ the \textbf{update rule} of $ A_{opt} $.  Let's define two characteristics of an update rule: The update rule, $A^t(.)$ is 
	$\mathbf{\sigma}$\textbf{-bounded} 
	if
	\begin{equation}\label{bounded}
	\sup_{\theta \in H} \left\| \theta-A^t(\theta) \right\| \leq \sigma,
	\end{equation} 
	and it is $ \mathbf{\tau\text{\textbf{-expensive}}} $ if	
	\begin{equation}\label{expensive}
	\sup_{\theta,\, \theta' \in H} \frac{\left\| A^t(\theta) - A^t(\theta') \right\|}{\left\| \theta - \theta' \right\|} \leq \tau,
	\end{equation}
	where $ \left\|.\right\| $ is the $L_2$ norm.	
	\begin{Lemma}[Growth recursion] \textbf{\cite{hardt2016train}} \label{GRR} 
		Given two training set $S$ and $S'$, suppose that $\theta_0, \theta_1, \ldots, \theta_T$ and $\theta'_0, \theta'_1,\ldots \theta'_T$ are two updates of network parameters with update rules $ A^t_S $ and
		$ A^t_{S'} $, running on $S$ and $S'$ respectively such that for each $ 1 \leq t \leq T $, $ \theta_{t} = A^t_S(\theta_{t-1}) $ and $ \theta'_{t} = A^t_{S'}(\theta'_{t-1}) $. For $\Delta_t = \left\| \theta_t - \theta'_{t} \right\|$, we have:
		\begin{itemize}
			\item
			If $A^t_S$ and $A^t_{S'}$ are equal and $\tau$-expensive, 
			then
			$\Delta_{t} \leq \tau \Delta_{t-1}$.
			\item
			If $A^t_S$ and $A^t_{S'}$ are $ \sigma\text{-bounded}$,
			then
			$\Delta_{t} \leq \Delta_{t-1} + 2\sigma $ \footnote{In the work of \cite{hardt2016train}, this inequality has been written as $\Delta_{t} \leq \min(1,\tau)\Delta_{t-1} + 2\sigma $ whose right side is less than $ \Delta_{t-1} + 2\sigma $ that we just need in the proofs of the theorems.}.
		\end{itemize}
	\end{Lemma}
	We state the proof of Lemma \ref{GRR} in Appendix \ref{appendix}. In Subsection \ref{adam}, we discuss the uniform stability of Adam to upper-bound the generalization error of a DNN trained by it. Subsequently, In Subsection \ref{adamw}, we state different theorems for the uniform stability of AdamW and the generalization error because AdamW exploits decoupled weight decay, and its update parameters statement is different from Adam.
	\subsection{Adam Optimizer} \label{adam}
	Let $ \ell(f^\theta; B) $ represents the computation of a loss function on an arbitrary mini-batch, $ B = \lbrace (\mathrm{x}_i, \mathrm{y}_i)\rbrace_{i=1}^{b} $,  which we use at each iteration to update parameters in order to minimize $ R_{emp}(f^\theta) $:
	\[  
	\ell(f^\theta; B) = \frac{1}{b}\sum_{i = 1}^{b}\ell(f^\theta(\mathrm{x}_i), \mathrm{y}_i), 
	\]
	in which $\theta$ are the parameters and $ b $ is the batch size. Let $ g(\theta) = \nabla_\theta \ell(f^\theta; B) $ where $ \nabla_\theta $ is the gradient. For $t\geq1$ suppose that  $ m_t $, $ v_t $  are estimates of the first and second moments respectively:
	\begin{align}
	m_t &= \beta_1 \cdot m_{t-1} + (1-\beta_1) \cdot g(\theta_{t-1}); \,\, m_0=0, \label{m_t} \\
	v_t &= \beta_2 \cdot v_{t-1} + (1-\beta_2) \cdot g^2(\theta_{t-1}); \,\, v_0=0 \label{v_t},
	\end{align}
	where $ \beta_1, \beta_2 \in (0,1)$ are exponential decay rates and the multiply operation is element-wise. Let $ \widehat{m}_t = m_t/(1-\beta_1^t) $ and $\widehat{v}_t = v_t/(1-\beta_2^t)$ be the bias-corrected estimates; Adam computes the parameters update using $ \widehat{m}_t $ adapted by $ \widehat{v}_t $:
	\[
	\theta_t = \theta_{t-1} - \eta \cdot \frac{\widehat{m}_t}{(\sqrt{\widehat{v}_t}+\epsilon)},
	\]
	where $ \eta $ is the learning rate and $ \epsilon=10^{-8} $. Based on what we discussed so far, to evaluate the uniform stability of Adam, we need to formulate its update rule. Given $\beta_1,\beta_2 \in (0,1)$ for each $ 1 \leq t \leq T $ let  
	\begin{align}
	\hat{M}(m_{t-1},\theta) &= \frac{\beta_1 \cdot m_{t-1} + (1-\beta_1) \cdot g(\theta)}{1-\beta_1^t}, \label{firstordereq}\\
	\hat{V}(v_{t-1},\theta) &= \frac{\beta_2 \cdot v_{t-1} + (1-\beta_2) \cdot g^2(\theta)}{1-\beta_2^t}, \label{secondordereq}
	\end{align}
	where $m_{t-1}$ and $v_{t-1}$ are the biased estimates for the first and second moments of the gradient at the previous step respectively as we explained in the equations \eqref{m_t} and \eqref{v_t}. Adam's update rule is obtained as follows:
	
	\begin{equation}\label{adamrule}
	A^t(\theta) = \theta - \eta \cdot \left(\frac{\hat{M}(m_{t-1},\theta)}{\sqrt{\hat{V}(v_{t-1},\theta)}+ \epsilon}\right),
	\end{equation}
	where $\eta$ is the learning rate and the division operation is element-wise. We use the following lemma in the proof of Theorem \ref{thm1}:
	\begin{Lemma}\label{firstMomentlemma}
		Let $ m_{t-1} = \beta_1 \cdot m_{t-2} + (1-\beta_1) \cdot g(\theta_{t-2}) $  such that $\beta_1 \in (0,1)$ is constant and $ m_0= 0 $. Let $\ell(\hat{\mathrm{y}}, \mathrm{y})$ be $ \gamma $-Lipschitz. Then for all $ t \geq 1 $ and $ \theta \in H$, we have $\left\| \hat{M}(m_{t-1},\theta) \right\| \leq \gamma$.
	\end{Lemma}
	The proof of Lemma \ref{firstMomentlemma} is available in Appendix \ref{appendix}. Now we can state the theorems which link the generalization error with the loss function properties. In Theorem \ref{thm1} we assess the stability measures including the uniform stability and in Theorem \ref{thm2}, we drive an upper bound for the generalization error of a DNN trained by Adam.
	 
	\begin{Theorem} \label{thm1}
		Assume Adam is executed for $ T $ iterations with a learning rate $ \eta $ and batch size $b$ to minimize the empirical risk in order to obtain $ f_{B_S,R} $. Let $\ell(\hat{\mathrm{y}}, \mathrm{y})$ be convex and $ \gamma $-Lipschitz. Then, Adam is $\beta$-uniformly stable with regard to the loss function $ \ell $, and for each $ (\mathrm{x},\mathrm{y}) $, $ \ell(f_{B_S,R}(\mathrm{x}),\mathrm{y}) $ holds the $\rho$-BDC with respect to $R$. Consequently, we have
		\[
		\beta \leq \frac{2\eta}{c} \cdot \frac{bT\gamma^2}{N}, \ \ \ \rho \leq \frac{8\eta}{c} \cdot \left( \frac{b\gamma}{N} \right)^2,
		\]
		in which $c \in (0,1)$ is a constant number and $ N $ is the size of the training set.
	\end{Theorem}
	\begin{proof}
		Consider Adam's update rule, $A^t(.)$ in the equation \eqref{adamrule}. In order to prove that $A^t(.)$ satisfies the conditions of Lemma \ref{GRR}, $ \sigma\text{-boundedness} $  and $\tau\text{-expensiveness} $ of $A^t(.)$ are needed to be evaluated. From the formula \eqref{bounded}, we have:
		\[
		\left\| \theta-A^t(\theta) \right\| = \left\| \eta \cdot \left( \frac{\hat{M}(m_{t-1},\theta)}{\sqrt{\hat{V}( v_{t-1},\theta)}+\epsilon} \right) \right\|,
		\] 
		where $m_{t-1}$ and $v_{t-1}$ are the biased estimates for $\mathbb{E}\left[g\right]$ and $\mathbb{E}\left[g^2\right] \geq 0$ in the $ t $-th step respectively. Therefore:
		\begin{align}
		\left\| \eta \cdot \left( \frac{\hat{M}(m_{t-1},\theta)}{\sqrt{\hat{V}( v_{t-1},\theta)}+\epsilon} \right) \right\|
		&\leq \eta \cdot \left\| \frac{\hat{M}(m_{t-1},\theta)}{\epsilon} \right\| \label{eq2} \\
		&\leq \frac{\eta\gamma}{\epsilon}. \label{eq3}
		\end{align}
		Because $ \epsilon > 0 $ and $ \hat{V}(v_{t-1},\theta) \geq 0$, we deduced the inequality \eqref{eq2}. In the inequality \eqref{eq3}, Lemma \ref{firstMomentlemma} has been applied, which implies that, $A^t(.)$ is $\sigma$-bounded such that $\sigma \leq \frac{\eta\gamma}{\epsilon}$. Now, we check the $ \tau\text{-expensiveness} $ condition:
		we know that for all $\theta \in H$, 	$ \frac{\hat{M}(m_{t-1},\theta)}{\sqrt{\hat{V}(v_{t-1},\theta)}} \simeq \pm 1 $ because $|\mathbb{E}[g]|/\sqrt{\mathbb{E}[g^2]} \leq 1$.
		On the other hand $\ell(\hat{\mathrm{y}}, \mathrm{y})$ 	is convex. Thus, for two updates of network parameters $\theta_{t-1}$ and $\theta'_{t-1}$ in an arbitrary iteration $t$ with the same initialization, by choosing a sufficiently small learning rate, the two vectors $ \frac{\hat{M}(m_{t-1},\theta_{t-1})}{\sqrt{\hat{V}(v_{t-1} ,\theta_{t-1})}} $
		and $ \frac{\hat{M}(m_{t-1},\theta'_{t-1})}{\sqrt{\hat{V}(v_{t-1},\theta'_{t-1})}} $
		are approximately equal. Thus, by substituting $A^t(.)$ in the formula \eqref{expensive}, it is concluded that, $A^t(.)$ is $1\text{-expensive}$. Note that in the training process, we only work with $ \theta_{t-1}, \theta'_{t-1} $ at each timestep $ t $. Therefore, according to the definition of $\Delta_{t-1}$ and the proof of the first case of Lemma \ref{GRR}, it is enough to prove $1\text{-expensiveness}$ for $ \theta_{t-1}, \theta'_{t-1} $ at each timestep $ t $.
		
		Let $B_S$ and $B_{S'}$ having equal size $k$, be two partitions of training sets $S$ and $S'$ respectively, such that $B_S$ and $B_{S'}$ are different in only one mini-batch. Let $\theta_0, \theta_1, \ldots, \theta_T$ and $\theta'_0, \theta'_1,\ldots, \theta'_T$ be two parameters updates obtained from training the network by Adam with update rules $ A^{t}_S $ and $ A^{t}_{S'} $ respectively where $ A^{t}_S $ runs on $ B_S $ and $A^{t}_{S'}$ runs on $ B_{S'} $ with the same random sequence $ R $ such that $\theta_0 = \theta'_0$. Let two mini-batches $ B $ and $ B' $ have been selected for updating the parameters in the $ t $-th iteration. If $ B=B' $, then $ A^t_{S'}=A^t_S $ else $ A^t_{S'} \neq A^t_S $.
		$B=B'$ occurs 	with probability $ 1-\frac1k $ and the opposite occurs with probability $ \frac1k $. At the beginning of the proof, we demonstrated that $A^t(.)$ (for an arbitrary training set) is
		$\sigma\text{-bounded}$ and $ 1 $-expensive. Let $ \Delta_t = \left\| \theta_t - \theta'_{t} \right\| $, from Lemma \ref{GRR}, we have:
		\begin{align*}
		\Delta_{t} &\leq (1-\frac1k)\Delta_{t-1} + \frac1k\left(\Delta_{t-1} +  \frac{2\eta\gamma}{\epsilon}\right)\\
		&= \Delta_{t-1} + \frac1k \cdot \frac{2\eta\gamma}{\epsilon}. 
		\end{align*}
		We know $ k=\frac{N}{b} $. Therefore, solving the recursive relation gives
		\[
		\Delta_T \leq \Delta_0 + 2T\eta \cdot \frac{\gamma}{k\epsilon} =
		2\eta \cdot \frac{bT\gamma}{N\epsilon}.
		\]
		Let $ \theta_{T,i} $ are the effective parameters of $\theta_T$ on the $ i $-th neuron of the last layer with $ M $ neurons. notation $ \langle .,. \rangle $ is inner product and
		$ \left[ f(i) \right]_{i=1}^{M} $ for an arbitrary function $f$, denotes the vector $[f(1), f(2),\ldots,f(M)]$. Now we proceed to prove Adam's uniform stability. According to Definition \ref{uniformstabiliydef}, we have:
		\begin{align}
		&\mathbb{E}_R \left( |\ell(f_{B_{S'},R}(\mathrm{x}),\mathrm{y}) - \ell(f_{B_S,R}(\mathrm{x}),\mathrm{y})| \right) \notag \\
		&\leq \mathbb{E}_R \left( \gamma \left\| f_{B_{S'},R}(\mathrm{x}) - f_{B_S,R}(\mathrm{x}) \right\| \right) \notag \\
		&= \gamma \mathbb{E}_R \left( \left\| \left[ \langle \theta'_{T,i}, \mathrm{x} \rangle \right]_{i=1}^{M} - \left[ \langle \theta_{T,i},\mathrm{x} \rangle \right]_{i=1}^{M} \right\| \right) \notag \\
		&\leq \gamma \mathbb{E}_R \left( \left\| \theta'_T - \theta_T \right\| \right) \label{eq5} \\
		&= \gamma\mathbb{E}_R\left[ \Delta_T \right] \notag \\
		&\leq 2\eta \cdot \frac{bT\gamma^2}{N\epsilon}. \label{eq6}
		\end{align}
		In the inequality \eqref{eq5}, we assumed $\left\|  \mathrm{x} \right\| \leq 1$; that is the re-scaling technique that is common in deep learning. In the last inequality, $\epsilon$ is a constant number between 0 and 1. 
		
		After showing the relation between the uniform stability of Adam and the Lipschitz constant of the loss function, we evaluate the bounded difference condition for the loss function with respect to the random sequence and a fixed training set. Suppose that $R$ and $R'$ are two random sequences of batch indices to update the parameters in which only the location of two indices has been changed; that is if $R = \left( \ldots,i,\ldots,j,\dots \right)$ then $R' = \left( \ldots,j,\ldots,i,\dots \right) $. Without loss of generality, assume $1 \leq i \leq \frac{k}{2}$
		and $\frac{k}{2}+1 \leq j \leq k$. The probability of selecting two identical batches in the $t$-th iteration is $1-\frac{4}{Tk^2}$. Thus, two updates of neural network parameters as $ \theta^R_0,\theta^R_1,\ldots,\theta^R_T $ and $ \theta^{R'}_0, \theta^{R'}_1,\ldots,\theta^{R'}_T $ are made with the same initialization, $ \theta^R_0 =\theta^{R'}_0$. Let $ \Delta_t = \left\| \theta^{R}_t - \theta^{R'}_{t} \right\| $. From Lemma \ref{GRR}, we have:
		\[
		\Delta_T \leq \frac{8}{Tk^2} \cdot \frac{\eta T\gamma}{\epsilon} = \frac{8}{k^2} \cdot \frac{\eta\gamma}{\epsilon}.
		\]
		According to Definition \ref{BDC}, we have:
		\begin{align}
		&|\ell(f_{B_S,R'}(\mathrm{x}),\mathrm{y}) - \ell(f_{B_S,R}(\mathrm{x}),\mathrm{y})| \notag \\ 
		&\leq \gamma \left\| f_{B_S,R'}(\mathrm{x}) - f_{B_S,R}(\mathrm{x}) \right\| \notag \\
		&= \gamma \left\| \left[ \langle \theta^{R'}_{T,i}, \mathrm{x} \rangle \right]_{i=1}^{M} - \left[ \langle \theta^R_{T,i},\mathrm{x} \rangle \right]_{i=1}^{M} \right\| \notag \\
		&\leq \gamma  \left\| \theta^{R'}_T - \theta^R_T \right\| \label{eq7} \\
		&= \gamma\Delta_T \notag \\
		&\leq \frac{8}{k^2} \cdot \frac{\eta \gamma^2}{\epsilon}.  \label{eq8}
		\end{align}
		The inequality \eqref{eq7} has been obtained similar to \eqref{eq6}. Replacing $k$ by $\frac{N}{b}$ in the inequality \eqref{eq8} leads to the inequality in the proposition.
	\end{proof}
	\begin{Theorem} \label{thm2}
		Let	$ \ell(\hat{\mathrm{y}}, \mathrm{y}) $ with the maximum value of $ L $ be convex and $ \gamma $-Lipschitz. Assume Adam is run for $ T $ iterations with a learning rate $ \eta $ and batch size $b$ to obtain $f_{B_S,R}$. Then we have the following upper bound for $E(f_{B_S,R})$ with probability at least $1-\delta $: 
		\begin{equation} \label{thm2-eq1}
		E(f_{B_S,R}) \leq \frac{2\eta}{c} \left( 4\left( \frac{b\gamma}{N} \right)^2 \sqrt{T\ log(2/\delta)} + \frac{bT\gamma^2}{N} \left( 1+\sqrt{2N\log(2/\delta)} \right) \right) + L\sqrt{\frac{\log(2\delta)}{2N}},
		\end{equation}
		in which $c \in (0,1)$ is a constant number and $ N $ is the size of the training set.
	\end{Theorem}
	\begin{proof}
		In the work of \cite{akbari2021does}, an upper bound for the generalization error of the output model trained by any optimization algorithm $A_{opt}$  is established with probability at least $1-\delta$, under the condition $A_{opt}$ satisfies uniform stability measure with bound $\beta$ and for each $ (\mathrm{x},\mathrm{y}) $, $ \ell(f_{B_S,R}(\mathrm{x}),\mathrm{y}) $ holds the $\rho$-BDC with regard to $R$ \footnote{In the assumptions of the main theorem in the work of \cite{akbari2021does}, it has been stated that the model is trained by stochastic gradient descent, but by studying the proof, we realize that their argument can be extended to any iterative algorithm that is $\beta$-uniformly stable because, in their proof, the upper bound has been derived independently of the update rule of stochastic gradient descent. The proof is available at http://proceedings.mlr.press/v139/akbari21a/akbari21a-supp.pdf.}:
		\begin{equation}\label{generalizationbound}
		E(f_{B_S,R}) \leq \rho\sqrt{T\log(2/\delta)} + \beta(1+\sqrt{2N\log(2/\delta)}) + L\sqrt{\frac{\log(2/\delta)}{2N}}. 
		\end{equation}  
		By combining Theorem \ref{thm1} and the inequality \eqref{generalizationbound}, we have the following upper bound with probability $1-\delta$:
		\begin{equation} \label{thm2-eq2}
		E(f_{B_S,R}) \leq \frac{2\eta}{c} \left( 4\left( \frac{b\gamma}{N} \right)^2 \sqrt{T\ log(2/\delta)} + \frac{bT\gamma^2}{N} \left( 1+\sqrt{2N\log(2/\delta)} \right) \right) + L\sqrt{\frac{\log(2\delta)}{2N}},
		\end{equation}
		where $c \in (0,1)$ is a constant number.
	\end{proof}
	Theorem \ref{thm2} shows how the generalization error bound of deep learning models trained by Adam depends on the Lipschitz constant $\gamma$ and the maximum value $L$. Furthermore, the inequality \eqref{thm2-eq1}, implies the sensitivity of the generalization error to the batch size; when the batch size grows, $ E(f_{B_S,R}) $ increases. On the other hand, from the basics of machine learning, we know, if the batch size is too small, the parameters update is very noisy. Thus, an appropriate value should be considered for the batch size according to the training set size.  
	
	As we mentioned in Section \ref{problem} for the KL and GJM losses we have $\gamma_{KL} \leq \gamma_{GJM}$ and $L_{KL} \leq L_{GJM}$ \cite{akbari2021does}. Hence, following Theorem \ref{thm2} we have the following corollary:
	\begin{Corollary} \label{corollary1}
		Let $ f^{KL}_{B_S,R} $ and $f^{GJM}_{B_S,R}$ be the output models trained by Adam optimizer, under the same settings using the KL and GJM loss functions respectively and the partition $B_S$ obtained from the training set $ S $. Then the upper bound of $ E(f^{GJM}_{B_S,R}) $ is less than the upper bound of $ E(f^{KL}_{B_S,R}) $.
		
	\end{Corollary}
	\begin{proof}
		We know if $\alpha=0.5$, then the Lipschitz constant and the maximum value of GJM are less than the Lipschitz constant and the maximum value of KL respectively \cite{akbari2021does}. So, under the same settings for hyper-parameters of Adam and the same initialization, from Theorem \ref{thm2}, the proposition is concluded.
		
	\end{proof}
	\subsection{AdamW Optimizer} \label{adamw}
	The objective of regularization techniques is to control the network parameters' domain in order to prevent the over-fitting issue. $L_2$-regularization which exploits the $L_2$ norm of the parameters vector, is more practical than $L_1$ because it keeps the loss function differentiable and convex. In continuation, we study $L_2$-regularization and note its effect on SGD and Adam. The lack of significant effect of this technique on Adam led to AdamW \footnote{Adam with decoupled weight decay} \cite{loshchilov2017decoupled}.
	
	Let $\ell^{reg}(f^\theta; B)$ be a regularized loss function computed on a mini-batch, $ B=\lbrace (\mathrm{x}_i, \mathrm{y}_i)\rbrace_{i=1}^{b} $:
	\begin{equation} \label{regloss}
	\ell^{reg}(f^\theta;B) = \frac1b\left( \sum_{i=1}^{b} \ell(f^\theta(\mathrm{x}_i), \mathrm{y}_i) + \frac{\lambda}{2}\left\| \theta \right\|^2 \right),
	\end{equation}
	where $\left\| . \right\|$ is the $L_2$ norm, $\lambda \in \mathbb{R^+}$ is the weight decay and $b$ is the batch size. According to the equation \eqref{regloss}, to compute the parameters update in SGD, we have: 
	\[ 
	\theta_{t} = \left(1-\frac{\eta\lambda}{b}\right)\theta_{t-1} - \frac{\eta}{b}\sum_{i=1}^{b} \nabla_{\theta} \ell(f^\theta(\mathrm{x}_i), \mathrm{y}_i).
	\]
	In SGD, minimizing the regularized loss function can improve the output model generalization. However, this technique, cannot be effective in Adam because it uses adaptive gradients to update the parameters \cite{loshchilov2017decoupled}. In AdamW, the weight decay hyper-parameter was decoupled from optimization steps taking gradient of the loss function. Let $\widehat{m}_t$ and $\widehat{v}_t$ denote the bias-corrected estimates illustrated in Subsection \ref{adam}. The parameters update is computed as follows:
	 \begin{equation} \label{paramupdateadamw}
	  \theta_t = \theta_{t-1} - \alpha_t \left( \eta \cdot \frac{\widehat{m}_t}{(\sqrt{\widehat{v}_t}+\epsilon)} + \lambda\theta \right),
	  \end{equation}
	  where $\alpha_t$ is the schedule multiplier. The equation \eqref{paramupdateadamw} exhibits that AdamW updates the parameters in a different way than Adam. Hence, we need to state theorems specific to AdamW for the stability and the generalization error.  Consider $\hat{M}(m_{t-1},\theta)$ and $\hat{V}(v_{t-1},\theta)$ in the equations \eqref{firstordereq} and \eqref{secondordereq}. According to parameters update statement of AdamW in formula \eqref{paramupdateadamw}, The AdamW's update rule is defined as
	  \begin{equation}\label{adamWrule}
	  A_W^t(\theta) = \theta - \alpha_t\left( \eta \cdot \frac{\hat{M}(m_{t-1},\theta)}{\sqrt{\hat{V}(v_{t-1},\theta)} + \epsilon} +\lambda \theta \right).
	  \end{equation}
	  where 0 < $\alpha_t\lambda < 1$ because otherwise, the update occurs in a wrong direction which means it goes away from the minimum. Consider $H$ as the set of all possible values for the network parameters. As we noted in Section \ref{prelim}, $H$ is bounded. Hence, the supremum of $H$ is well-defined. Let  $ \left\| \theta \right\|_{\sup} \vcentcolon =   \sup_{\theta \in H} \left\| \theta \right\| $:
	  
	  	\begin{Theorem} \label{thm3}
	  	Assume AdamW is executed for $ T $ iterations with a learning rate $ \eta $, batch size $b$, weight decay $\lambda$, and schedule multiplier $\alpha_t$ to minimize the empirical risk in order to obtain $ f_{B_S,R} $. Let $\ell(\hat{\mathrm{y}}, \mathrm{y})$ be convex and $ \gamma $-Lipschitz. Then, Adam is $\beta$-uniformly stable with regard to the loss function $ \ell $, and for each $ (\mathrm{x},\mathrm{y}) $, $ \ell(f_{B_S,R}(\mathrm{x}),\mathrm{y}) $ holds the $\rho$-BDC with respect to $R$. Consequently, we have
	  	\[
	  	\beta \leq \frac{2bT}{N} \sum_{t=1}^{T} \alpha_t \left( \frac{\eta\gamma^2}{c} + \gamma \lambda \left\| \theta \right\|_{\sup} \right), \ \ \ \rho \leq \frac{8b^2}{N^2} \sum_{t=1}^{T} \alpha_t \left( \frac{\eta\gamma^2}{c} + \gamma \lambda \left\| \theta \right\|_{\sup} \right),
	  	\]
	  	in which $c \in (0,1)$ is a constant number and $ N $ is the size of the training set.
	  \end{Theorem}
  		\begin{proof}
  		First, we check the $\sigma$-boundedness of $A_W^t(\theta)$:
  		\begin{align}
  		\left\| \theta - A_W^t(\theta) \right\| &= \left\| \alpha_t \left( \eta \cdot \frac{\hat{M}(m_{t-1},\theta)}{\sqrt{\hat{ V}(v_t,\theta)} + \epsilon} + \lambda\theta \right) \right\| \notag \\
  		&\leq \left\|\alpha_t\eta \cdot \frac{\hat{M}(m_{t-1},\theta)}{\epsilon} \right\| + \alpha_t\lambda \left\| \theta \right\| \notag \\
  		&= \alpha_t \left( \eta \cdot \frac{\left\| \hat{M}(m_{t-1},\theta)\right\|}{\epsilon} + \lambda\left\| \theta \right\| \right)
  		\notag \\
  		& \leq \alpha_t \left(\frac{\eta\gamma}{\epsilon} + \lambda\left\| \theta \right\|_{\sup} \right). \label{sigmaneq}
  		\end{align}
  		By applying Lemma \ref{firstMomentlemma}, we concluded the inequality \eqref{sigmaneq}, which shows that $A_W^t(\theta)$ is $\sigma$-bounded. Now we evaluate the $\tau$-expensiveness of AdamW. According to the formula \eqref{expensive}, we have
  		\begin{align}
  		&\frac{\left\| A_W^t(\theta) - A_W^t(\theta') \right\|}{\left\|\theta-\theta' \right\|} \notag \\
  		&= \frac{\left\| -\alpha_t \left( \eta \cdot \frac{\hat{M}(m_{t-1},\theta)}{\sqrt{\hat{V}(v_{t-1},\theta)} + \epsilon} + \lambda\theta \right) + \alpha_t \left( \eta \cdot \frac{\hat{M}(m_{t-1},\theta')}{\sqrt{\hat{V}(v_{t-1},\theta')} + \epsilon} + \lambda\theta' \right) + \theta - \theta' \right\|}{\left\| \theta-\theta'\right\|}. \label{eq14}
  		\end{align}
  		As said in the proof of Theorem \ref{thm1}, for every $\theta \in H$, we have $\frac{\hat{M}(m_{t-1},\theta)}{\sqrt{\hat{V}(v_{t-1},\theta)}} \simeq \pm 1$ because $|\mathbb{E}[g]|/\sqrt{\mathbb{E}[g^2]} \leq 1$. Therefore, the equation 
  		\eqref{eq14} is written as follows:
  		\begin{align}
  		\frac{\left\| -\alpha_t\lambda\theta + \alpha_t\lambda\theta' + \theta - \theta' \right\|}{\left\| \theta - \theta' \right\|} &= \frac{\left\| \alpha_t\lambda(\theta'-\theta) + \theta - \theta' \right\|}{\left\| \theta - \theta' \right\|} \notag \\
  		&= \frac{|1-\alpha_t\lambda|\left\| \theta-\theta' \right\|}{\left\| \theta-\theta' \right\|} \notag \\
  		&= |1-\alpha_t\lambda| < 1. \label{eq15}
  		\end{align}
  		AdamW update rule in the equation \eqref{adamWrule} implies that $ 0 < \alpha_t\lambda < 1 $ which its consequent is the inequality \eqref{eq15}. with an analogous demonstration to what we did in the proof of Theorem \ref{thm1}, i.e. considering update sequences and using Lemma \ref{GRR}
  		in order to evaluate the uniform stability and bounded difference condition according to their definitions, we conclude the following inequalities:
  		\begin{align*}
  		&\beta \leq \frac{2bT}{N} \sum_{t=1}^{T} \alpha_t \left( \frac{\eta\gamma^2}{\epsilon} + \gamma \lambda \left\| \theta \right\|_{\sup} \right),\\ 
  		&\rho \leq \frac{8b^2}{N^2} \sum_{t=1}^{T} \alpha_t \left( \frac{\eta\gamma^2}{\epsilon} + \gamma \lambda \left\| \theta \right\|_{\sup} \right).
  		\end{align*}
  	\end{proof}
  	\begin{Theorem} \label{thm4}
  		Let $ \ell(\hat{\mathrm{y}}, \mathrm{y}) $ with the maximum value of $ L $ be convex and $ \gamma $-Lipschitz. Assume AdamW is run for $ T $ iterations with a learning rate $ \eta $, batch size $b$, weight decay $\lambda$, and schedule multiplier $\alpha_t$ to obtain $f_{B_S,R}$. Then we have the following upper bound for $E(f_{B_S,R})$ with probability at least $1-\delta $: 
  		\begin{equation} \label{thm4-eq1}
  		E(f_{B_S,R}) \leq \frac{2b}{N} \sum_{t=1}^{T} \alpha_t \left( \frac{\eta\gamma^2}{c} + \gamma \lambda \left\| \theta \right\|_{\sup} \right) \left( \frac{4b}{N}\sqrt{T\log(2/\delta)} + T\sqrt{2N\log(2/\delta)} \right) + L\sqrt{\frac{\log(2/\delta)}{2N}}.
  		\end{equation}
  	in which $c \in (0,1)$ is a constant number and $ N $ is the size of the training set.
  \end{Theorem}
	\begin{proof}
	By combining the equation \eqref{generalizationbound} and Theorem \ref{thm3} we conclude the proposition.
	\end{proof}
	The inequality \eqref{thm4-eq1} implies that the generalization error growth of a DNN trained by AdamW, is directly related to the Lipschitz constant and the maximum value of a loss function. Following Theorem \ref{thm4} we have the following corollary for the KL and GJM loss functions:
	
	\begin{Corollary} \label{corollary2}
		Let $ f^{KL}_{B_S,R} $ and $f^{GJM}_{B_S,R}$ be the output models trained by AdamW optimizer, under the same settings using the KL and GJM loss functions respectively and the partition $B_S$ obtained from the training set $ S $. Then the upper bound of $ E(f^{GJM}_{B_S,R}) $ is less than the upper bound of $ E(f^{KL}_{B_S,R}) $.
	\end{Corollary}
	\begin{proof}
		The proposition is concluded by Theorem \ref{thm4} and an analogous argument to Corollary \ref{corollary1}.
	\end{proof}
	\section{Experimental Evaluation}
	\subsection{Datasets}
	We use 4 datasets, including UTKFace \cite{zhang2017age}, AgeDB \cite{moschoglou2017agedb}, MegaAge-Asian \cite{huang2016unsupervised}, and FG-NET \cite{chen2013cumulative} to evaluate age estimation performance. UTKFace dataset contains $ 23,708 $ facial images, providing enough samples of all ages, ranging from 0 to 116 years-old. AgeDB contains $16,488$ in-the-wild images in the age range from $0$ to $100$ years-old. MegaAge-Asian has been already split into MegaAge-Train and Mega-Age-Test datasets, containing $40,000$ and $3,945$ images respectively, belonging to Asian people with the age label in the range from $1$ to $69$ years-old. FG-NET dataset contains $1,002$ facial images in the age range of $0$ to $69$ years. This dataset covers variations in pose, age expression, resolution,  and lighting conditions. By collecting the samples from  UTKFace, MegaAge-Train, and AgeDB datasets whose ages are in the range from $ 0 $ to $ 100 $ years-old, we create a new dataset called UAM, which includes $80,174$ images. We use UTKFace and UAM  as the training sets. FG-NET, MegaAge-Test, and 10\% randomly selected from AgeDB called AgeDB-Test, are left as the test sets.
	\subsection{Settings}
	All images are pre-processed by the following procedures: face detection and alignment are done by prepared modules in OpenCV package. All images are reshaped to the size of $256 \times 256$ and standard data augmentation techniques, including random cropping and horizontal flipping, are carried out during the training phase. 	We use two neural network architectures VGG16 \cite{vgg16} and ResNet50 \cite{resnet50}, pre-trained on ImageNet \cite{deng2009imagenet} and VGGFace2 \cite{cao2018vggface2} datasets respectively, to estimate human age. VGGFace2 dataset was created with the aim of estimating human pose and age. With the same seed, the last layer of these models is replaced with a $ \mathrm{M} $-neurons dense layer with random weights. The last layer of VGG16 is trained on UTKFace in 5 epochs and the last layer of ResNet50 is trained on UAM in 15 epochs. $ \mathrm{M} $ is set to $ 116 $ in VGG16 and $101$ in ResNet50 model. We train the models via Adam and AdamW with learning rate $ 2 \times 10^{-5}$  for KL and $10^{-4}$ for GJM \footnote{In our experiments, when we set the learning rate to $2 \times 10^{-5}$ for the GJM loss, the ultimate model at the last epoch remained under-fit.}. The batch size and AdamW's weight decay are set to 64 and 0.1 respectively. We set $\beta_1=0.9$ and $\beta_2=0.999$ for both Adam and AdamW as the authors of \cite{kingma2014adam} and \cite{loshchilov2017decoupled} suggested.
	
	\subsection{Evaluation Metrics and Results}
\begin{figure}
		\begin{center}
			\begin{subfigure}[b]{0.32\textwidth}
				\includegraphics[height=3.33cm, width=5cm]{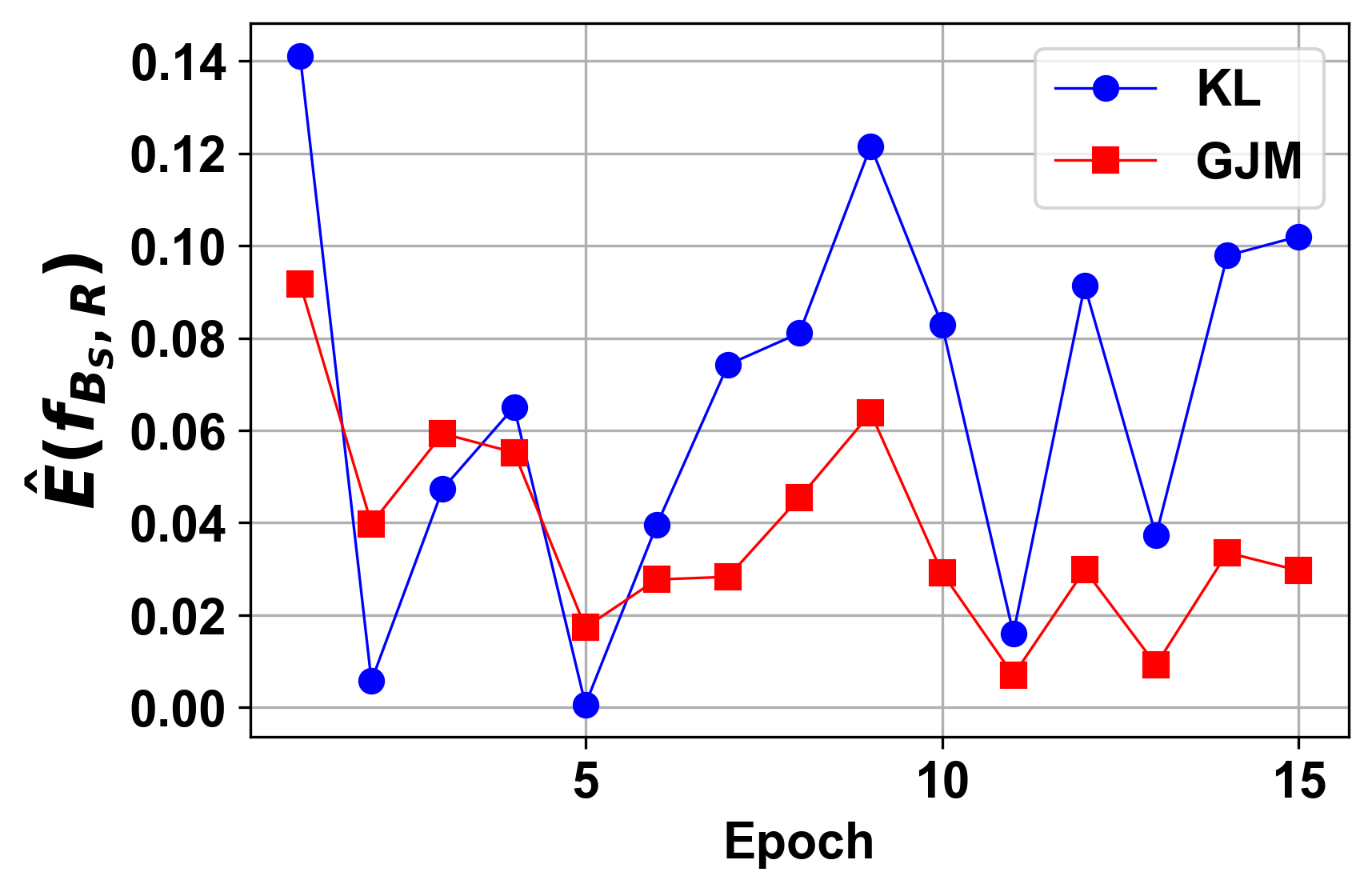}
				\subcaption{Estimate of the generalization error}
				\label{GE-ResNet-Adam}
			\end{subfigure}
			\hfill
			\begin{subfigure}[b]{0.32\textwidth}
				\includegraphics[height=3.33cm, width=5cm]{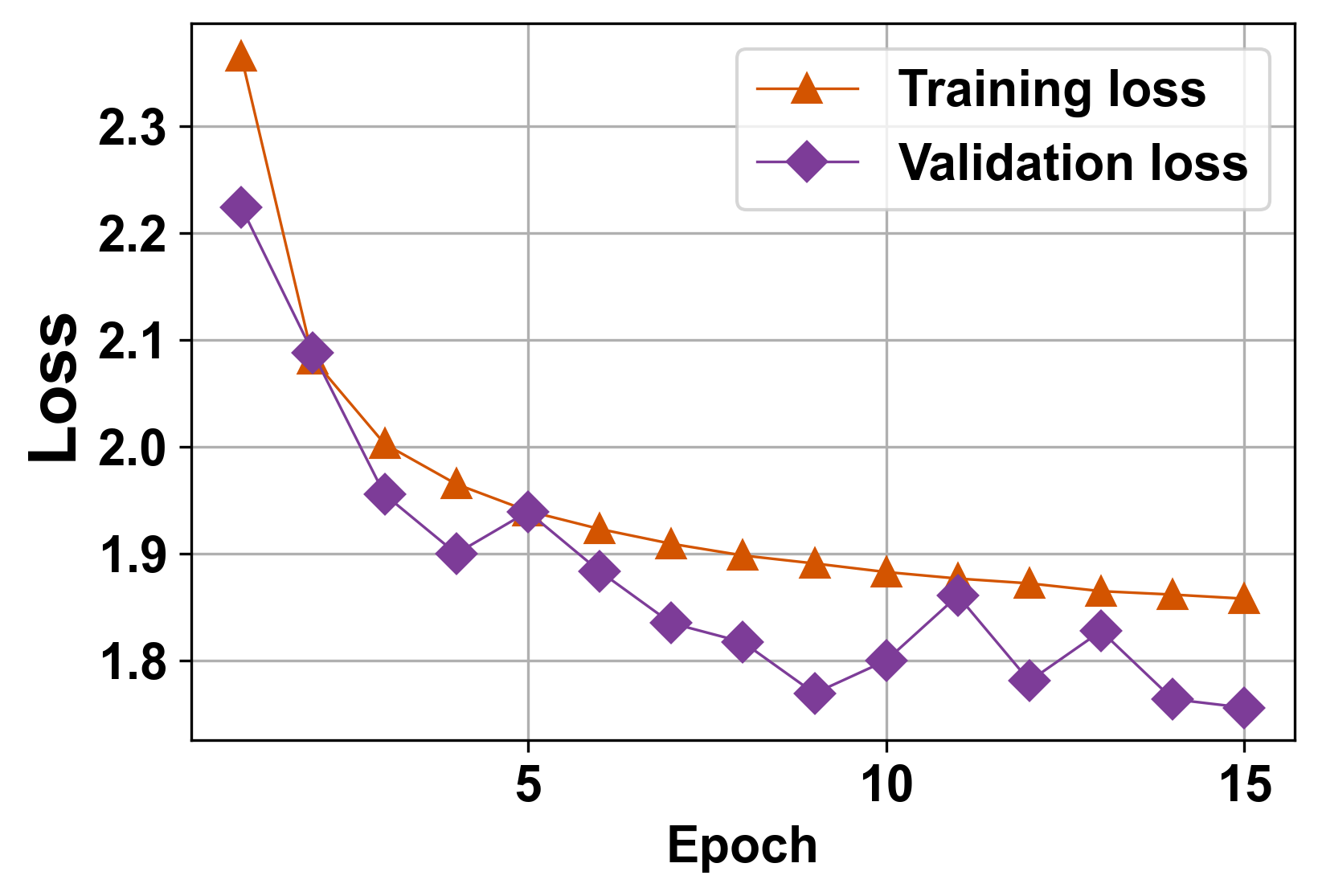}
				\subcaption{Training and validation loss of KL}
				\label{kl-loss-ResNet-Adam}
			\end{subfigure}
			\hfill
			\begin{subfigure}[b]{0.32\textwidth}
				\includegraphics[height=3.33cm, width=5cm]{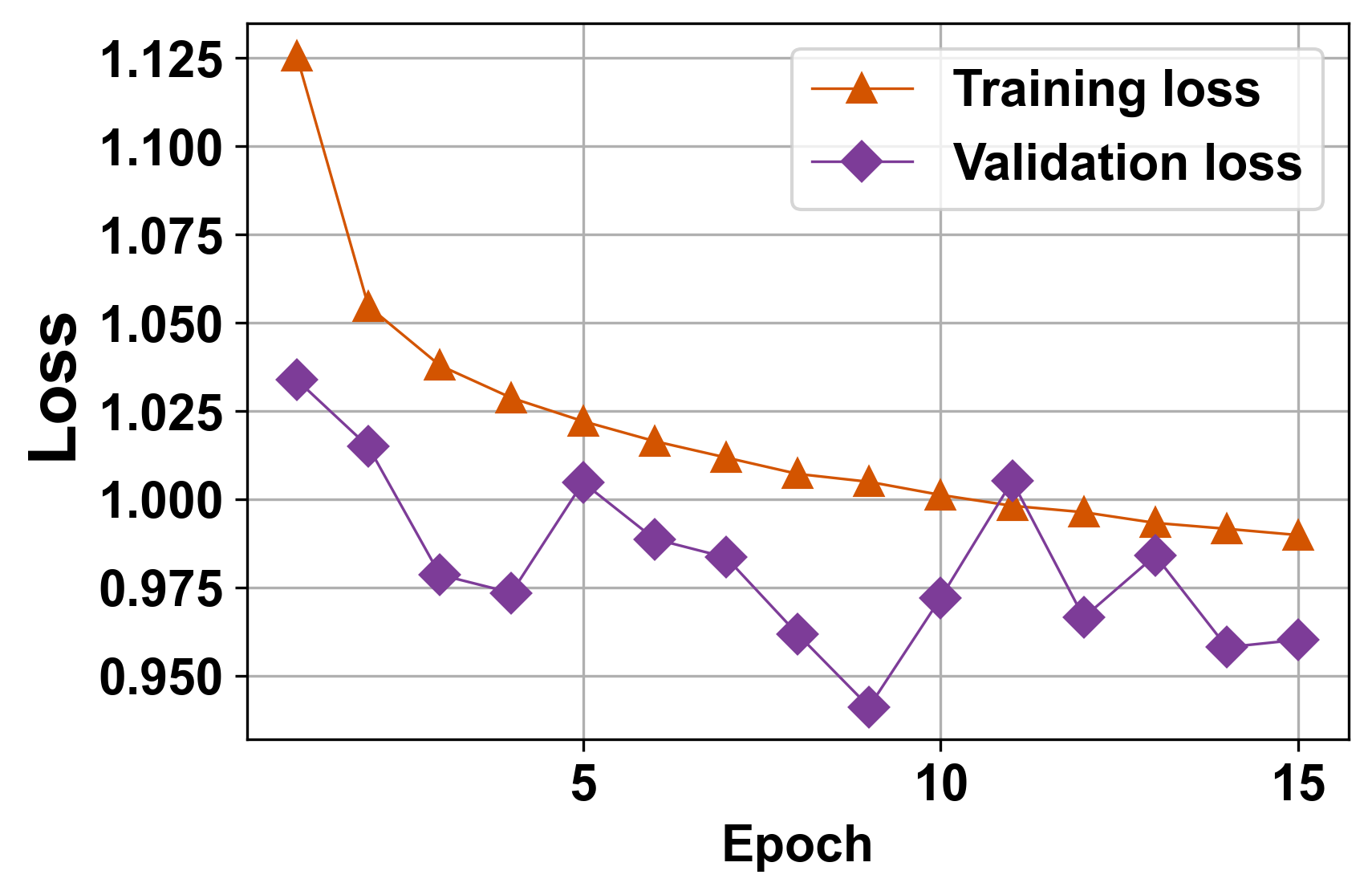}
				\subcaption{Training and validation loss of GJM}
				\label{gjm-loss-ResNet-Adam}
			\end{subfigure}
		\caption{Loss measurement (Model: ResNet50, Optimzer: Adam, Training set: UAM, Validation set: FG-NET)}
		\label{fig1}
		\end{center}
	\end{figure}

\begin{figure}
		\begin{center}
			\begin{subfigure}[b]{0.32\textwidth}
				\includegraphics[height=3.33cm, width=5cm]{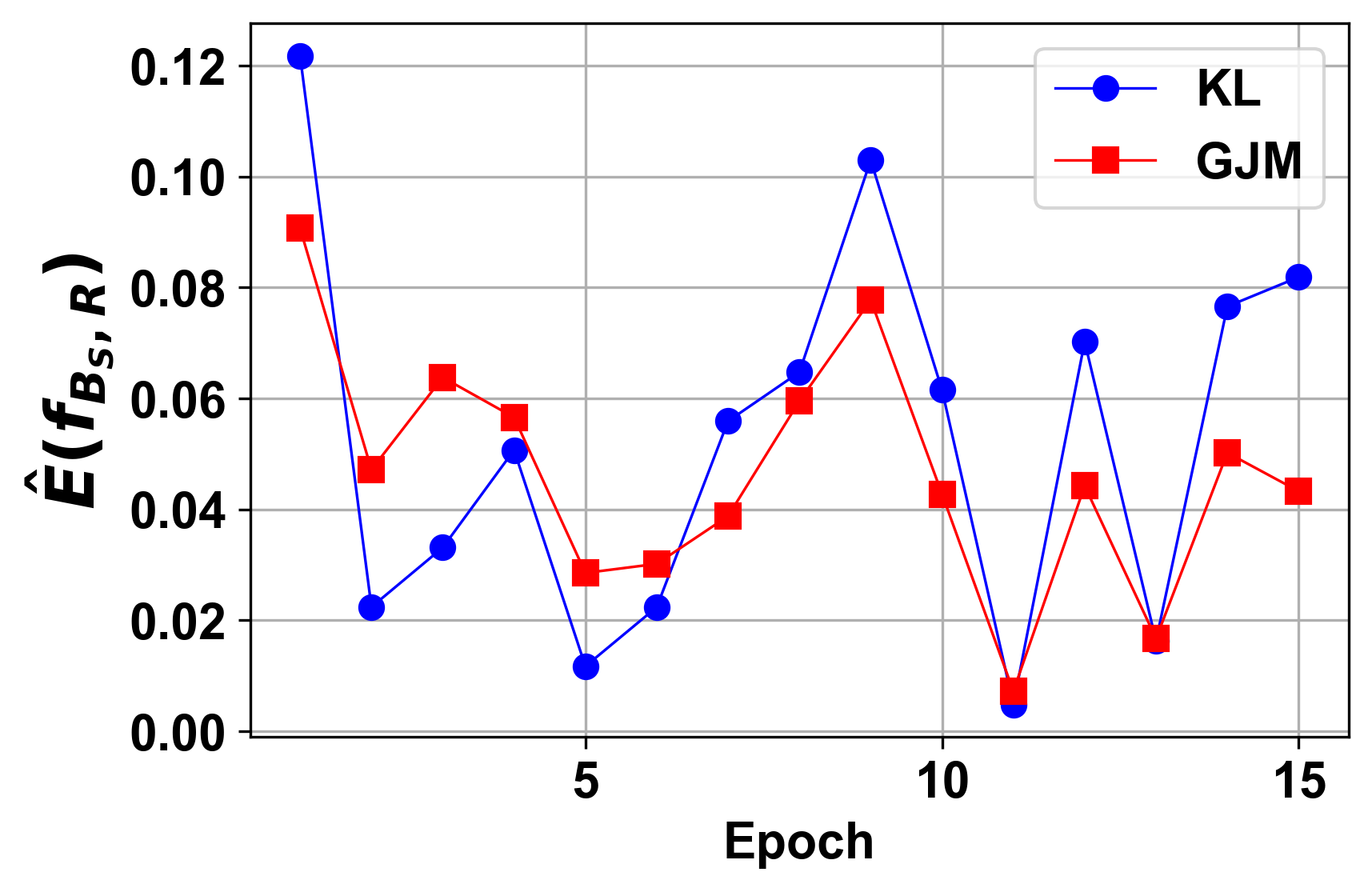}
				\subcaption{Estimate of the generalization error}
				\label{GE-ResNet-Adamw}
			\end{subfigure}
			\hfill
			\begin{subfigure}[b]{0.32\textwidth}
				\includegraphics[height=3.33cm, width=5cm]{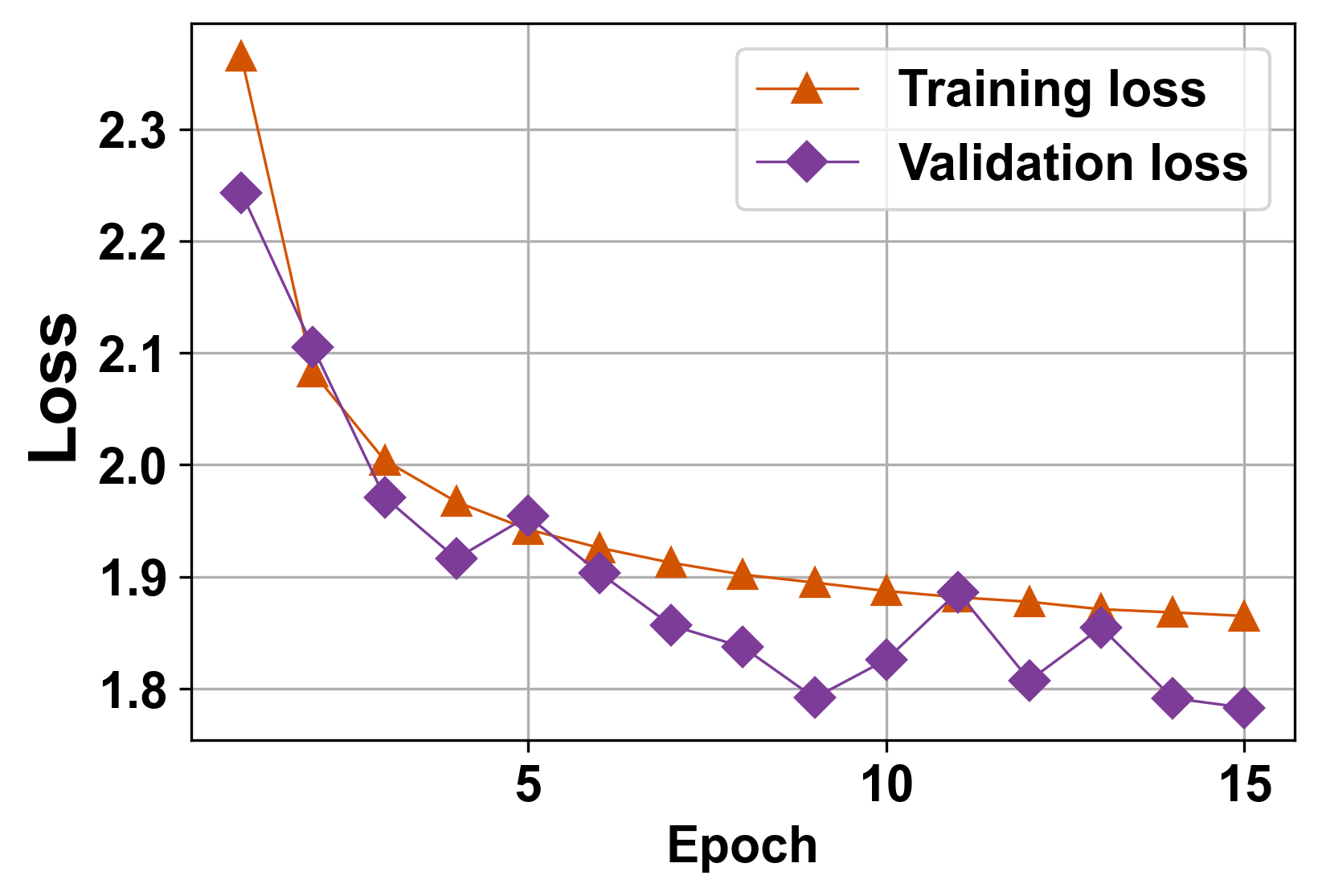}
				\subcaption{training and validation loss of KL}
				\label{kl-loss-ResNet-Adamw}
			\end{subfigure}
			\hfill
			\begin{subfigure}[b]{0.32\textwidth}
				\includegraphics[height=3.33cm, width=5cm]{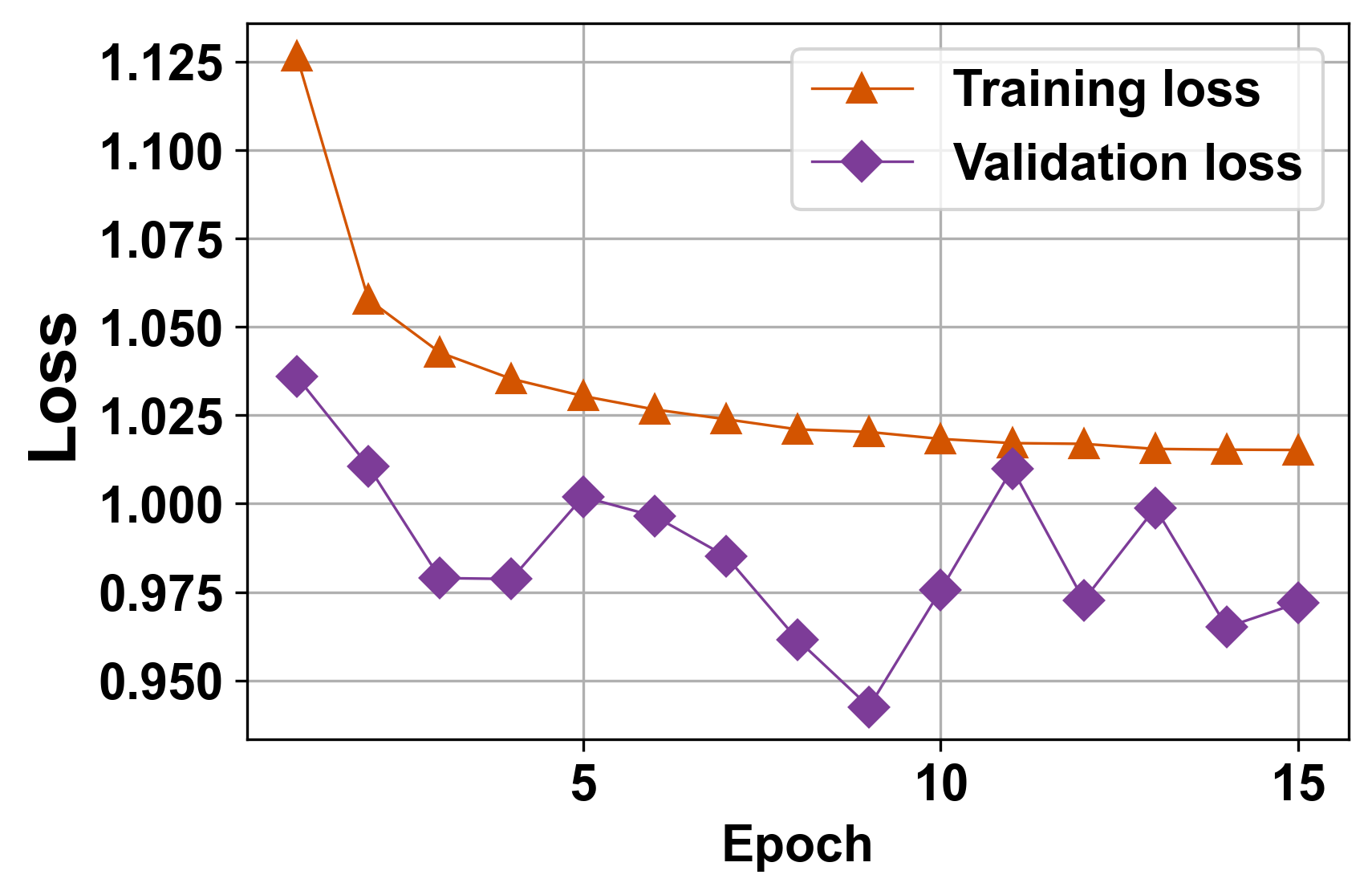}
				\subcaption{Training and validation loss of GJM}
				\label{gjm-loss-ResNet-Adamw}
			\end{subfigure}
		\caption{Loss measurement (Model: ResNet50, Optimzer: AdamW, Training set: UAM, Validation set: FG-NET)}
		\label{fig2}
		\end{center}
	\end{figure}

	As the first observation, we measure the generalization error estimate in the training steps of ResNet50 trained by Adam and AdamW which is defined as
	\[\hat{E}(f_{B_S,R}) = |R_{train} (f_{B_S,R}) - R_{val} (f_{B_S,R}) |, \]
	where $f_{B_S,R}$ is the output model, $ R_{train} (f_{B_S,R}) $, $ R_{val} (f_{B_S,R}) $ are the average of loss values on the training and validation sets respectively. The results of this experiment are shown in Figure \ref{fig1} and Figure \ref{fig2}. In the first epochs, the models are still under-fit and the loss is far from its minimum; therefore, $\hat{E}(f_{B_S,R})$ does not give us critical information about the generalization error, but in the rest of epochs, when the experimental loss of the models approaches its minimum, $ \hat{E}(f_{B_S,R}) $ can represent the generalization error. As can be seen in Figure \ref{GE-ResNet-Adam} and Figure \ref{GE-ResNet-Adamw}, after epoch 5 or 6 the generalization error estimate of the models trained by Adam and AdamW using the GJM loss function is lower than the models trained using the KL loss.
	
	 In addition, we measure the generalization performance in terms of Mean Absolute Error (MAE) and Cumulative Score (CS). Consider the training set $S$, and the test set $S_{test} \in (X \times Y)^D$. Let $(\mathrm{x}_k, y_k) \in S_{test}$ represents a test example where $ y_k \in \mathbb{N} $ is the label of $ k $-th example of the test set. Since we use label distribution learning, for each $(\mathrm{x}, \mathrm{y}) \in S$, $ \mathrm{y} \in \mathbb{R}^\mathrm{M}$ is the probability distribution corresponding to $\mathrm{x}$. Therefore, in the evaluation phase, the output of the model per the test example $ \mathrm{x}_k $ is the predicted probability distribution $\hat{\mathrm{y}}_k = \left[\hat{y}_{k,1},\hat{y}_{k,2}, \ldots, \hat{y}_{k ,\mathrm{M}} \right]$. MAE is defined as  $\frac{1}{D}\sum_{k=1}^{D} | \hat{l}_k - l_k |$ where $\hat{l}_k$ is the index of the largest element of $ \hat{\mathrm{y}}_k $ and $ l_k $ is the true label. CS  is defined as $\frac{D_I}{D} \times 100\%$ where $D_I$ is the number of test samples such that $|\hat{l}_k - l_k| < I$. Commonly, the value of $I$ is set to $5$ \cite{akbari2021does}\cite{Shen_2018_CVPR}. 
	 
	 The results are reported in Tables \ref{vgg16-adamw-utkface}-\ref{resnet50-adam-uam}. The ResNet50 models are more accurate than the VGG16 models because VGG16 is pre-trained on ImageNet dataset which is not suitable for age estimation. Tables \ref{vgg16-adamw-utkface}- \ref{resnet50-adam-uam} show that when we train a DNN by Adam or AdamW, the GJM loss performs better than the KL loss.
	
	\begin{table}[h]
		\begin{center}
			\caption{MAE and CS metrics (Model: VGG16, Optimizer: AdamW, Training set: UTKFace)}
			\label{vgg16-adamw-utkface}
			\begin{tabular}{|l|cc|cc|}
				\hline
				& \multicolumn{2}{c|}{FG-NET} &  \multicolumn{2}{c|}{AgeDB-Test} \\
				\hline
				\hline
				Method & MAE &  CS(\%) & MAE  & CS(\%) \\
				\hline
				LDL (KL)  & $ 14.41 $ & $ 15.67 $ & $ 22.77 $ & $ 16.13 $ \\
				LDL (GJM) & $ \textbf{14.11} $ & $ \textbf{17.06} $ & $ \textbf{22.53} $ & $ \textbf{17.04} $ \\
				\hline
			\end{tabular}
		\end{center}
		
		\begin{center}
			\caption{MAE and CS metrics (Model: ResNet50, Optimizer: AdamW, Training set: UAM)}
			\label{resnet50-adamw-uam}
			\begin{tabular}{|l|cc|cc|}
				\hline
				& \multicolumn{2}{c|}{FG-NET} &  \multicolumn{2}{c|}{MegaAge-Test} \\
				\hline
				\hline
				Method & MAE &  CS(\%) & MAE  & CS(\%) \\
				\hline
				 LDL (KL) & $ 5.55 $ & $ 54.49 $ & $ 6.23 $ & $ 50.59 $ \\	
				 LDL (GJM) & $ \textbf{5.48} $ & $ \textbf{54.90} $ & $ \textbf{6.21} $ & $ \textbf{51.58} $ \\
				\hline
			\end{tabular}
		\end{center}

		\begin{center}
			\caption{MAE and CS metrics (Model: ResNet50, Optimizer: Adam, Training set: UAM)}
			\label{resnet50-adam-uam}
			\begin{tabular}{|l|cc|cc|}
				\hline
				& \multicolumn{2}{c|}{FG-NET} &  \multicolumn{2}{c|}{MegaAge-Test} \\
				\hline
				\hline
				Method & MAE &  CS(\%) & MAE  & CS(\%) \\
				\hline
				LDL (KL) & $ 5.43 $ & $ 55.09 $ & $ 6.28 $ & $ 49.96 $ \\
				LDL (GJM) & $ \textbf{5.27} $ & $ \textbf{56.57} $ & $ \textbf{6.16} $ & $ \textbf{51.91} $ \\
				\hline
			\end{tabular}
		\end{center}
	\end{table}

	\section{Conclusion}
	In this paper, we have shown how the properties of a loss function affect the generalization performance of a DNN trained by Adam or AdamW. We theoretically linked the Lipschitz constant and the maximum value of a loss function to the generalization error of the output model obtained by Adam or AdamW. We evaluated our theoretical results in the human age estimation problem, where we trained and tested the models on various datasets. In our future work, we focus on the tasks addressed with single-label and multi-label learning instead of label distribution learning because there are several classification tasks in computer vision, recommender systems, predictive modeling, etc where the models do not need to learn the label distribution, and the training process is done using the cross-entropy loss. Our next step is proposing an alternative loss function to cross-entropy to improve the generalization performance of single-label and multi-label learning models. 
	
	
	\bibliographystyle{unsrt}
	\bibliography{BibFile}
	
	\appendix 
	\numberwithin{equation}{section}
	\section{Appendix: Proofs of Lemmas} \label{appendix}
	\subsection{Proof of Lemma \ref{GRR}}
	\begin{proof}
		\begin{itemize}
		\item The first case:
		\[ \Delta_{t} = \left\| \theta_{t} - \theta'_{t} \right\| = \left\| A^t_{S}(\theta_{t-1}) - A^t_{S'}(\theta'_{t-1}) \right\| = \left\| A^t_S(\theta_{t-1}) - A^t_S(\theta'_{t-1}) \right\| \leq \tau \left\| \theta_{t-1}-\theta'_{t-1} \right\|. \]
		\item The second case:
		\begin{align*}
		\Delta_{t} &= \left\| \theta_{t} - \theta'_{t} \right\| \\
		&= \left\| A^t_{S}(\theta_{t-1}) - A^t_{S'}(\theta'_{t-1}) \right\| \\
		&= \left\| (A^t_{S'}(\theta'_{t-1}) - \theta'_{t-1}) + (\theta_{t-1} - A^t_S(\theta_{t-1})) + \theta'_{t-1} - \theta_{t-1} \right\|\\
		&\leq \left\| \theta'_{t-1} - \theta_{t-1} \right\| + \left\| A^t_{S'}(\theta'_{t-1})-\theta'_{t-1} \right\| + \left\| \theta_{t-1} - A^t_S(\theta_{t-1}) \right\| \\
		&\leq \Delta_{t-1} + 2\sigma.
		\end{align*}
		\end{itemize}
	\end{proof}
	
	\subsection{Proof of Lemma \ref{firstMomentlemma}}
	\begin{proof}
		We know $ \left\| \nabla_{\theta} \ell(f^\theta(\mathrm{x}),\mathrm{y}) \right\| \leq \gamma$ for each $(\mathrm{x}, \mathrm{y})$. Hence, for every $\theta \in H$ and a mini-batch $B = \lbrace (\mathrm{x}_i, \mathrm{y}_i) \rbrace_{i=1}^{b}$ we have  
		\[\left\| g(\theta) \right\| = \left\| \nabla_\theta \ell(f^\theta; B) \right\| = \left\| \frac{1}{b}\sum_{i = 1}^{b} \nabla_\theta \ell(f^\theta(\mathrm{x}_i), \mathrm{y}_i) \right\| \leq \gamma. \] The proposition is clear for $t=1$. For $t > 1$  by solving the recurrence relation of $m_{t-1}$ we have:
		\[
		\hat{M}(m_{t-1},\theta) = \frac{(1-\beta_1) \sum_{j=1}^{t-1}\beta_1^{t-j}g(\theta_{j-1}) + (1-\beta_1)g(\theta)}{1-\beta^ t_1}.
		\]
		Now we proceed to draw the conclusion:
		\begin{align}
		\left\| \hat{M}(m_{t-1},\theta) \right\| &= \left\| \frac{(1-\beta_1) \sum_{j=1}^{t-1}\beta_1^{t-j}g(\theta_{j-1}) + (1-\beta_1)g(\theta)}{1-\beta^t_1} \right\| \notag \\
		&\leq \frac{(1-\beta_1) \sum_{j=1}^{t-1}\beta_1^{t-j} \left\| g(\theta_{j-1}) \right\| + (1-\beta_1)\left\| g(\theta) \right\|}{1-\beta^t_1} \notag \\
		&\leq \frac{(1-\beta_1)\sum_{j=1}^{t}\beta_1^{t-j}}{1-\beta^t_1} \cdot \gamma \notag \\
		&= \gamma. \notag
		\end{align}
	\end{proof}
	
\end{document}